\documentclass{article}

\usepackage{journal2e_v1}
\usepackage[a4paper]{geometry}

\usepackage{amssymb}
\usepackage{times}
\usepackage{helvet}
\usepackage{courier}
\usepackage{amsfonts}
\usepackage{amsmath}
\usepackage{booktabs}
\usepackage{graphicx}
\usepackage{subfigure}

\def\A{{\bf A}}
\def\B{{\bf B}}

\def\L{{\bf L}}
\def\W{{\bf W}}
\def\X{{\bf X}}

\def\XM{{\mathcal X}}
\def\RB{{\mathbb R}}
\def\vect{\operatorname{vec}}
\def\reshape{\operatorname{reshape}}
\def\conv{\operatorname{conv}}
\DeclareMathOperator*{\argmin}{arg min}

\begin{document}
%
\title{Exploiting Local Structures with the Kronecker Layer in Convolutional Networks}
\author{
\name Shuchang Zhou     \\
\addr
Megvii Inc. \\
and State Key Laboratory of Computer Architecture \\
Institute of Computing Technology \\
Chinese Academy of Sciences, Beijing, China\\
\texttt{shuchang.zhou@gmail.com} \\
          \AND
\name Jia-Nan Wu \\
 \addr Megvii Inc. \\
              \texttt{wjn@megvii.com}
\AND
\name Yuxin Wu \\
 \addr Megvii Inc. \\
and Robotics Institute \\
Carnegie Mellon University \\
\texttt{ppwwyyxx@gmail.com}
\AND
\name Xinyu Zhou     \\
\addr Megvii Inc. \\
\texttt{zxy@megvii.com}
}

\maketitle
\begin{abstract}
\begin{quote}
In this paper, we propose and study a technique to reduce the number of parameters and computation time in convolutional neural networks.
We use Kronecker product to exploit the local structures within convolution and fully-connected layers, by replacing the large weight matrices by combinations of multiple Kronecker products of smaller matrices.
Just as the Kronecker product is a generalization of the outer product from vectors to matrices, our method is a generalization of the low rank
approximation method for convolution neural networks. We also introduce combinations of different shapes of Kronecker product to increase modeling capacity.
Experiments on SVHN, scene text recognition and ImageNet dataset demonstrate that we can achieve $3.3 \times$ speedup or $3.6 \times$ parameter reduction with less than 1\% drop in accuracy, showing the effectiveness and efficiency of our method.
Moreover, the computation efficiency of Kronecker layer makes using larger feature map possible, which in turn enables us to outperform the previous state-of-the-art on both SVHN(digit recognition) and CASIA-HWDB (handwritten Chinese character recognition) datasets. 
\end{quote}
\end{abstract}
\section{Introduction}
Recently, convolutional neural networks (CNNs) have achieved a great success in many computer vision and machine learning tasks. This success facilitates the development of industrial applications using CNNs. However, there are two major challenges for practical use of these networks, especially on resource-limited devices:
\begin{enumerate}
\item Using a CNN for prediction may require significant amount of computation at run time. For example, AlexNet\cite{krizhevsky2012imagenet} would
  require billions of floating point operations to process an image of $227\times 227 $size.
\item CNNs achieving state of the art results may require billions \cite{dean2012large,le2013building,Jaderberg14c} of parameters for storage.
\end{enumerate}
As a consequence, there has been growing interest in model speedup and compression. It is common to sacrifice a little prediction accuracy in exchange
for smaller model size and faster running speed.

In the literature, a major technique is based on the idea of low rank matrix and tensor approximations.
In \cite{sainath2013low}, low rank matrix factorization was used on the weight matrix of the final softmax layer. \citeauthor{denil2013predicting} (\citeyear{denil2013predicting}) decomposed the weight matrix as a product of two smaller matrices and one of the matrices was carefully constructed as a dictionary. In \cite{xue2013restructuring,denton2014exploiting}, model approximation is followed by fine-tuning on the training data. \citeauthor{zhang2015accelerating} (\citeyear{zhang2015accelerating}) also took the nonlinear activation functions into account when doing approximation.

\begin{figure}
\centering
\subfigure[]{
\includegraphics[height=0.18 \linewidth, width=0.27 \linewidth]{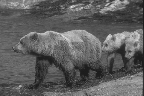}
}
\subfigure[]{
\includegraphics[height=0.72 \linewidth, width=0.27 \linewidth]{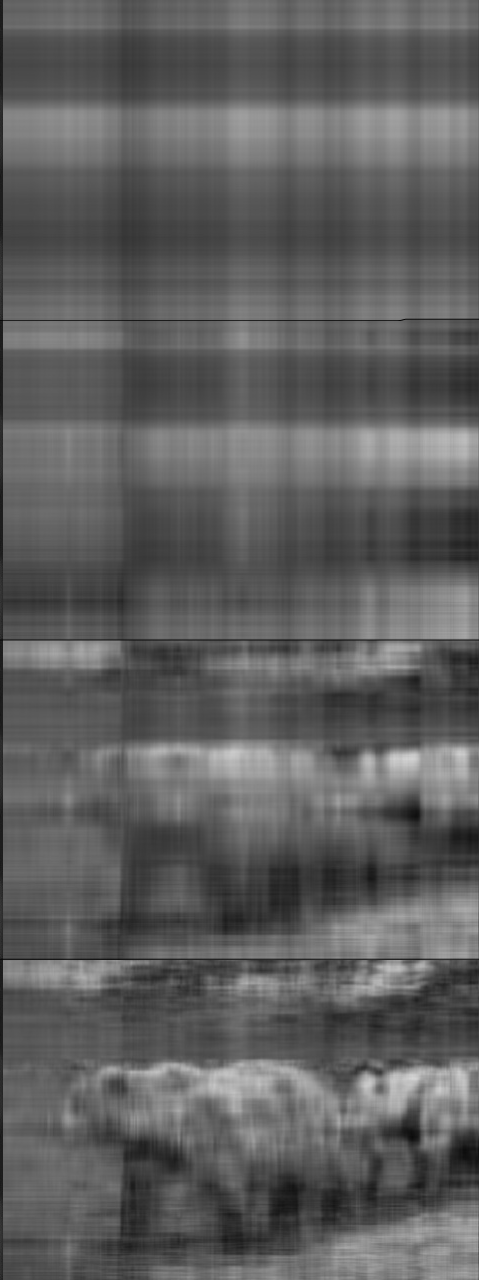}
}
\subfigure[]{
\includegraphics[height=0.72 \linewidth, width=0.27 \linewidth]{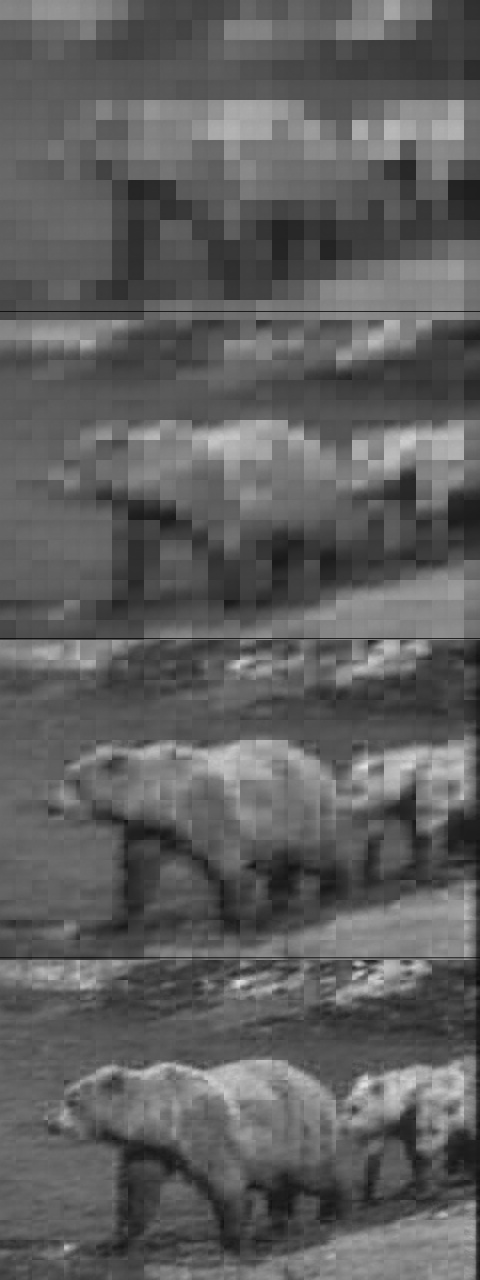}
}
\caption{Comparison between approximations by outer product and Kronecker product for an image. The column (a) is the origin image of size $480\times 320$, selected from BSD500 dataset \cite{amfm_pami2011}. The column (b) is the SVD approximations of (a) by outer product and the column (c) is the approximation based on Kronecker product\cite{van1993approximation}, with rank $1, 2, 5, 10$ respectively from top to down. The shape of the right matrix in the Kronecker product is deliberately selected as $20\times 16$ to make the number of parameters equal for each rank.}
\label{fig:bears}
\end{figure}

Low rank technique can also be applied on the weight tensors of convolutional layers. \citeauthor{rigamonti2013learning}
(\citeyear{rigamonti2013learning}) used a shared set of separable (rank-1) filters to approximate the original filters. \citeauthor{DBLP:conf/bmvc/JaderbergVZ14} (\citeyear{DBLP:conf/bmvc/JaderbergVZ14}) exploited the redundancy that exists between different feature channels and filters.  \citeauthor{lebedev2014speeding} (\citeyear{lebedev2014speeding}) applied CP-decomposition, a type of tensor decomposition, on the weight tensors.

In this paper, we explore a framework for approximating the weight matrices and weight tensors in neural networks by sum of Kronecker products.
 We note that as
 the bases for low rank factorizations like SVD or CP-decomposition are outer products of vectors, approximation by these bases can only exploit the redundancy \emph{along each dimension}. In contrast, as the Kronecker product generalizes the outer product from vectors to matrices of arbitrary shape, we may use the Kronecker product to exploit
 redundancy between local patches of \emph{any} shape.

 Figure \ref{fig:bears} demonstrates a case when approximating by Kronecker product would produce less reconstruction error than outer products with
 the same number of parameters for image pixel value matrix. Intuitively, similar situation may also exist for weight matrices and tensors in
 convolutional networks, and in these cases our method may produce approximate models that run faster and have less number of parameters at the same
 level of accuracy loss. On the other hand, with similar number of parameters, our method can advances previous state-of-the-art.

The rest of this paper is organized as follows. In Section 2, we introduce the Kronecker layer. We discuss some details about implementing the Kronecker layer in Section 3. We extend our technique to convolutional layer in Section 4. Section 5 analyses the result of using Kronecker layers on some benchmark datasets. Section 6 discusses some related work not yet covered. Finally, Section 7 concludes the paper and discusses future work.

\section{Kronecker Layer}
In this section, we first review the property of the Kronecker product and describe its application on the fully-connected layer.

\subsection{Kronecker Product}
Let $\A \in \RB^{m_1\times n_1}$, $\B\in \RB^{m_2\times n_2}$ be two given matrices. Then the Kronecker product $\A \otimes \B$
is an $m \times n$ matrix, where $m=m_1m_2, n=n_1n_2$:
\begin{eqnarray}
\A \otimes \B = \left[\begin{array}{ccc}
a_{11}\B & \cdots & a_{1n_1}\B \\
\vdots & \ddots & \vdots \\
a_{m_11}\B & \cdots & a_{m_1n_1}\B
\end{array} \right].
\end{eqnarray}
An important property with Kronecker product is that, it can be presented by matrix multiplication with reshape operation:
\begin{eqnarray}
\label{vectorization}
(\A\otimes\B)\vect(\X) = \vect(\B \X \A^{T}),
\end{eqnarray}
for  the matrix $\X \in \RB^{n_2 \times n_1}$. Here  $\vect(\X) =(x_{11}, \ldots, x_{n_2 1}, x_{12}, \ldots, x_{n_2 n_1})^T \in \RB^{n_2 n_1}$ denotes the vectorization (column vector) of the matrix $\X$. Below we will show how to speedup calculation of Kronecker products in neural networks using this property.

Kronecker products are easy to generalize from matrices to tensors. Let $\mathcal{A} \in \RB^{p_1 \times \cdots \times p_k}$ and $\mathcal{B} \in \RB^{q_1 \times \cdots \times q_k}$ and define:
\begin{align}
\begin{split}
&(\mathcal{A}\otimes\mathcal{B})_{i_1,\cdots,i_k} \\
=& \mathcal{A}_{\lfloor i_1/p_1 \rfloor, \cdots, \lfloor i_k/p_k \rfloor} \mathcal{B}_{i_1\bmod q_1, \cdots, i_k\bmod q_k},
\end{split}
\end{align}
where $\mathcal{A}\otimes\mathcal{B} \in \RB^{p_1q_1\times\cdots\times p_kq_k}$.

\subsection{Approximating the Fully-Connected Layer}
We next show how to use Kronecker products to approximate weight matrices of fully-connected layers, leading to construction of a new kind of layer
which we call a Kronecker fully-connected layer, or KFC layer for short. The idea originates from the observation that for a matrix $\W\in\RB^{m\times
  n}$ where the dimensions are not prime (in fact the dimension is commonly set to a multiple of 8, to make full use of modern CPU/GPU architecture)
we have approximation:
\begin{align}
\label{composition}
\W \approx \A \otimes \B,
\end{align}
where $m = m_1 m_2$, $n = n_1 n_2$, $\A\in \RB^{m_1\times n_1}$, $\B\in
\RB^{m_2\times n_2}$.
So the KFC layer is:
\begin{align}
\L_{i+1} = f((\A \otimes \B) \L_{i} + \bf{b}_i),
\end{align}
where $\L_i$ is the input of $i$th layer, $f$ is the nonlinear activation function.
With the case that $m$ or $n$ is prime, it suffices to add some extra dummy features or output classes to make the dimension a composite number.

\begin{figure}
\centering
\subfigure[]{
\includegraphics[height=0.2375 \linewidth, width=0.8 \linewidth]{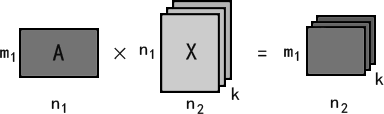}
}
\subfigure[]{
\includegraphics[height=0.219 \linewidth, width=0.8 \linewidth]{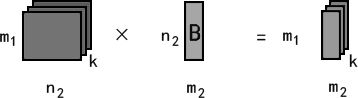}
}

\caption{A simple visualization of the computation procedure of the KFC layer. (a) denotes $\XM \times_1 \A$, (b) denotes the result of (a)$\times_2 \B$. The KFC layer transform a matrix from $(n_1n_2)\times k$ to $(m_1m_2)\times k$.}
\label{fig:fc}
\end{figure}

Note that we need not to calculate the Kronecker product $\A \otimes \B$ explicitly.
When the KFC layer is fed with inputs of batch $k$, we can forward the KFC layer efficiently, according to Eq. \eqref{vectorization}:
\begin{align}
\label{kp-tensor-calc}
\begin{split}
& (\A \otimes \B) [\vect{\X}_1, \cdots, \vect{\X}_k] \\
& =   [\vect
(\B \X_1 \A^\top), \cdots,
\vect(\B \X_k \A^\top)] \\
&=  \operatorname{Reshape}{(\XM \times_1 \A \times_2 \B)},
,
\end{split}
\end{align}
where $\XM \in\RB^{n_1\times n_2\times k}$ is a tensor stacked by $[\X_1^\top,\cdots,\X_k^\top]$, $k$ is the batch size. $\times_p$ is the tensor-matrix product over mode $p$ \cite{Kolda:2009:TDA:1655228.1655230}, which can be implemented as a matrix product following a linear time unfolding operation of tensor. The $\operatorname{Reshape}$ operator reshapes the tensor from $m_1 \times m_2 \times k$ to $m \times k$, which has nearly no overhead. Figure \ref{fig:fc} illustrates this procedure.
Similarly, the backward process is also simply matrix product.

Just as SVD approximation may be extended beyond rank-$1$ to an arbitrary rank, one could extend Kronecker product approximation to a sum of Kronecker products approximation. In addition, unlike the outer product, $\A$ and $\B$ may have different shapes. Hence, we get a more general KFC layer:
\begin{align}
\label{full_kpsvd}
\L_{i+1} = f\left(\left(\sum_{i=1}^{r} \A_i \otimes \B_i\right)\L_i + \bf{b}_i\right),
\end{align}
where $\A_i \in \RB^{m_1^{(i)}\times n_1^{(i)}}$, $\B_i^{(i)} \in \RB^{m_2^{(i)}\times n_2^{(i)}}$, $m_1^{(i)}m_2^{(i)}=m, n_1^{(i)}n_2^{(i)}=n$ and
$r \in \mathbb{N^+}$.

The number of parameters of a KFC layer is $\sum_{i=1}^{r}(m_1^{(i)} n_1^{(i)} + m_2^{(i)} n_2^{(i)})$ (bias terms are omitted),
reduced from $mn = m_1 n_1 m_2 n_2$. In particular, when all the small matrices have the same shape, the number is $r(m_1n_1+m_2n_2)$.

The computation complexity is $O(\sum_{i=1}^{r}(m_1^{(i)} n k+n_2^{(i)} m k))$, reduced from $O(mnk)$.  When all the small matrices have the same shape, it is $O(\frac{mnk}{n_1}+\frac{mnk}{m_2})$.

\begin{figure}
\centering
\includegraphics[height=0.4 \linewidth, width= \linewidth]{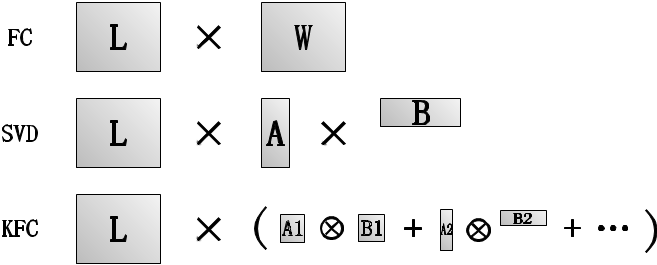}
\caption{Illustration of the fully-connected layer, fully-connected layer with SVD approximating and the KFC layer.}
\label{fig:kfc_compare}
\end{figure}

In particular, let $\A_i \in \RB^{1\times n_1}$, $\B_i \in \RB^{m_2 \times 1}$. The Kronecker product degenerates to the outer product, and the
approximation degenerates to a SVD method \cite{xue2013restructuring,denton2014exploiting}. Let $\A_i \in \RB^{m \times n}$, $\B_i \in \RB^{1 \times
  1}$. The KFC layer degenerates to the classical fully-connected layer. Figure \ref{fig:kfc_compare} illustrates the difference among
fully-connected layer, fully-connected layer with SVD approximation, and our KFC layer.

In the rest of the paper, we use the following notation to describe a configuration of a KFC layer as an approximation of
a FC layer with weight matrix $W \in \mathbb{R}^{m\times n}$: $ \{ (m_1^{(1)},m_2^{(1)}, n_1^{(1)}, n_2^{(1)}),\cdots ,(m_1^{(r)},m_2^{(r)},
n_1^{(r)},n_2^{(r)}) \}$ denotes a KFC layer of rank $r$,
with $ A_i \in \mathbb{R}^{m_1^{(i)}\times n_1^{(i)}}, B_i\in \mathbb{R}^{m_2^{(i)} \times n_2^{(i)}}, 1\le i \le r$.
In particular, we use the 5-tuple $ (m_1, m_2, n_1, n_2, r)$ to denote a KFC layer of rank $r$, where all components have the same shape.

\section{Details of the KFC Layer}
We now consider some details about the KFC layer, including how to initialize the layer, how to select the shapes and use more nonlinearity.

\subsubsection{Initialization}
KFC layers can be randomly initialized with the same method as with FC layers. However, in the case where we want to compress or speed-up a pre-trained model (for example, to run on mobile devices),
KFC layer can be initialized by approximating the pre-trained weight matrix $\W$, just like SVD method. The initialization problem can be formulated as the nearest KP problem.
\begin{align}
\argmin_{\A_i, \B_i} \|\W - \sum_{i=1}^{r}\A_i \otimes \B_i\|_F.
\end{align}
\citeauthor{van1993approximation}  (\citeyear{van1993approximation}) solved this problem with KPSVD when the shapes of $A_i, 0\leq i \leq r$, are the same. KPSVD bears strong connection with SVD. In fact, it can be turned into the following SVD decomposition using $\mathcal{R}$  operator:
\begin{align}
  \mathcal{R}(\W) = \sum_{i=1}^{r}\sigma_i \mathbf{u}_i \mathbf{v}_i^T,
\end{align}
where $\W\in \RB^{m\times n}$, $\mathcal{R}$ is a reordering operation and $\mathcal{R}(\W) \in \RB^{m_1n_1 \times m_2n_2}$.
$\mathbf{u}_i \in \RB^{m_1n_1}$ and $\mathbf{v}_i^T \in \RB^{m_2n_2}$. Then we have:
\begin{align}
  \vect\left({A_i}^{(opt)}\right) &= \sqrt{\sigma_i}\mathbf{u}_i, \\
  \vect\left({B_i}^{(opt)}\right) &= \sqrt{\sigma_i}\mathbf{v}_i.
\end{align}
For multiple shapes, we can apply KPSVD for one shape and reconstruct the weight matrix
$\tilde{W} = \sum_{i=1}^{r_s}A_i\otimes B_i$, where $r_s (\leq r)$ denotes the rank under certain shape. Then we apply KPSVD on $(W-\tilde{W})$ with the second shape. Repeat the above two steps recursively until all shapes are computed.

\subsubsection{Shape selection}
Any factors of $m$ and $n$ may be selected as $m_1$ and $n_1$ in the formula \ref{full_kpsvd}. However, in CNNs, the input to a fully-connected layer may be a tensor of order 4, namely, $\mathcal{L}_{i}\in \RB^{c \times h \times w \times k}$, where $c$ is the number of channels, $h$ is the height, $w$ is the  width and $k$ is the batch size. $\mathcal{L}_i$ is often reshaped into a matrix before being fed into a fully-connected layer as $\L_{i} \in \RB^{(chw) \times k}$. Though the reshaping transformation from $\mathcal{L}_{i}$ to $\L_{i}$ does not incur any loss in pixel values of data, we note that the dimension information is lost in the matrix representation. Due to the shape of $\W$, we may propose a few kinds of structural constraints by requiring $\W$ to be the Kronecker product of matrices of particular shapes.
\begin{itemize}
\item \textbf{Formulation \uppercase\expandafter{\romannumeral1}}: In this formulation, we require $n_1=c$ and $n_2=hw$. The number of parameters is reduced to $r(cm_1 + hwm_2)$. The underlying assumption for this model is that the channel transformation should be decoupled from the spatial transformation.
\item \textbf{Formulation \uppercase\expandafter{\romannumeral2}}: In this formulation, we require $n_1 = ch$ and $n_2=w$. The number of parameters is reduced to $r(chm_1 + wm_2)$. The underlying assumption for this model is that the transformation w.r.t. columns may be decoupled.
\item \textbf{Formulation \uppercase\expandafter{\romannumeral3}}: In this formulation, we require $n_1 = cw$,  $n_2=h$, and $\mathcal{L}_{i}$ needs to swap the second and the third dimension first. The number of parameters is reduced to $r(cwm_1 + hm_2)$. The underlying assumption for this model is that the transformation w.r.t. rows may be decoupled.
\end{itemize}
Of course, we can also combine the above three formulation together.

Otherwise, when the input is a matrix, we do not have natural choices of $m_1$ and $n_1$. Through experiments, we find it is possible to arbitrarily pick a decomposition of input matrix dimensions to enforce the Kronecker product structural constraint. It is also sensible to set $m_1n_1$ as close to $\sqrt{mn}$ as possible with a small $r$ to get a maximum compression ratio. But a smaller $m_2$ and $n_2$ and correspondingly larger $m_1$ and $n_1$ generally gives less accuracy loss. Nevertheless, we can use multiple components with different shapes to remedy the arbitrariness of the selection.

\subsubsection{Introducing More Nonlinearity}
When $r$ is not very large, we can move the summation out of the nonlinear function $f$ in Eq. \eqref{full_kpsvd} to introduce more nonlinearity to the KFC layer with little overhead:
\begin{align}
\begin{split}
\L_{i+1} & = \sum_{i=1}^{r} f(\left(\A_{i} \otimes \B_{i}\right) \L_{i} + \bf{b}_i),
\end{split}
\end{align}
The number of parameters only increases a little (more bias terms) or we can share the bias to avoid the increment. We have found the additional nonlinearity in the KFC layer  is very helpful sometimes.
Note the additional nonlinearity makes KFC layers difficult to be initialized by KPSVD. But it is not a serious problem. Initializing KFC layers with random number works well in our experiments.

\section{Generalization : the KConv Layer}
Since the fully-connected layer is a kind of $1\times 1$ convolution, we expand our work to the convolutional layer. In this section, we describe how to use Kronecker products to approximate weight tensors of convolutional layers, leading to construction of a new kind of layers which we call Kronecker convolutional layers, or KConv layers for short. We assume stride is $1$, no zero padding for simplicity in this section.
Weights of the convolutional layer can be described as a $4$-dimensional tensor: $\mathcal{W} \in \RB^{o\times c\times h \times w}$,
where $o$ is the number of output channels, $c$ is the number of input channels, $h$ and $w$ are the spatial dimensions of the kernel. The weight
tensor can be approximated as:
\begin{align}
\mathcal{W} \approx \sum_{i=1}^{r}\mathcal{A}_i\otimes\mathcal{B}_i,
\end{align}
with $\mathcal{A}_i \in \RB^{o_1^{(i)}\times c_1^{(i)}\times h_1^{(i)}\times w_1^{(i)}}$, $\mathcal{B}_i^{(i)} \in \RB^{o_2^{(i)}\times c_2^{(i)}\times h_2^{(i)}\times w_2^{(2)}}$, $c_1^{(i)}c_2^{(i)}=c$, $o_1^{(i)}o_2^{(i)}=o$ and $r \in \mathbb{N}$, $r > 0$. The shapes of the filters are constrained: $h_1^{(i)}h_2^{(i)}=h, h_1^{(i)}+h_2^{(i)}-1=h$;  $w_1^{(i)}w_2^{(i)}=w, w_1^{(i)}+w_2^{(i)}-1=w$. These constraints are the same as two schemes discussed in \cite{DBLP:conf/bmvc/JaderbergVZ14}.

Similar to the KFC layer, we do not need to calculate the tensor Kronecker product explicitly. For each shape $c_1,c_2,o_1,o_2$, we can replace the
original convolutional layer with two consecutive convolutional layers. Here we use rank $r=1$ for simplicity. The input is denoted as $\mathcal{L}_i \in \RB^{k\times c\times x \times y}$, where $x$,$y$ are the height and width, $k$ is the batch size. The KConv layer proceeds as following:
\begin{enumerate}
\item $\mathcal{L}_i^{'} = \reshape(\mathcal{L}_i) \in \RB^{kc_2\times c_1\times x \times y}$.
\item $\mathcal{L}_i^{''} = \conv(\mathcal{L}_i^{'}, \mathcal{A}) \in \RB^{kc_2\times o_1\times (x-h_1+1) \times (y-w_1+1)}$.
\item $\mathcal{L}_i^{'''} = \reshape(\mathcal{L}_i^{''}) \in \RB^{ko_1\times c_2\times (x-h_1+1) \times (y-w_1+1)}$.
\item $ \mathcal{L}_i^{''''} = \conv(\mathcal{L}_i^{'''}, \mathcal{B}) \in \RB^{ko_1\times o_2\times (x-h+1) \times (y-w+1)}$.
\item $ \mathcal{L}_{i+1} \mathrel{{+}{=}} \reshape({\mathcal{L}_i^{''''}}) \in \RB^{k\times o_1o_2\times (x-h+1) \times (y-w+1)}$.
\end{enumerate}
\begin{figure}
\centering
\includegraphics[height=0.71 \linewidth, width= \linewidth]{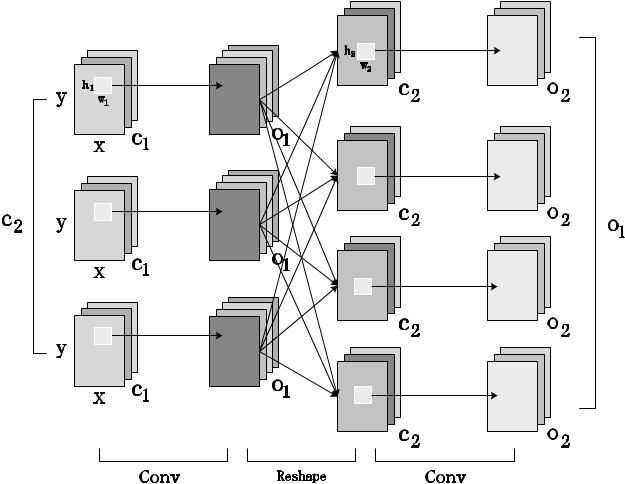}
\caption{KConv Layer can be implemented via a sequence of two convolutional layers with some reshape operation.}
\label{fig:kconv}
\end{figure}

Figure \ref{fig:kconv} illustrates the KConv framework. The number of parameters reduces to $o_1c_1h_1w_1+\frac{ochw}{o_1c_1h_1w_1}$ from $ochw$. The computation complexity reduces to $O(r(k(x-h+1)(y-w+1)(\frac{oc}{o_2}h_1w_1+\frac{oc}{c_1}h_2w_2)))$ from $O(k(x-h+1)(y-w+1)ochw)$.

In particular, if $o_1=1, c_1=c, h_1=h, w_1=1$, the KConv is the same as  Scheme 1 in \cite{DBLP:conf/bmvc/JaderbergVZ14}.
If $o_1=1, c_1=c, h_1=1, w_1=1$, the KConv is the same as Scheme 2 in \cite{DBLP:conf/bmvc/JaderbergVZ14}.
It is worth mentioning that with a rank $ r>1$ and $h_1=w_1=1$, the two convolution frameworks are very similar to the inception
\cite{szegedy2014going}, where the main difference is that inception has an extra $3\times 3$ max pooling branch.
However, KConv will allow more choices of $o_1$ and $c_1$.

\section{Experiments}
In this section, we empirically study the properties and efficiency of the Kronecker layer and compare it with some other common low rank model
approximation methods. As is well known, a large proportion of parameters in a CNN are contained in the fully-connected layers and most computation time is spent in the convolutional layer. Therefore, in the experiments, we mainly consider model acceleration in the convolutional layer and model compression in the fully-connected layer.

To make a fair comparison, for each dataset, we train a convolutional neural network as a baseline.
Then we replace the convolutional layer or fully-connected layer with a KConv or KFC layer and train the new network until quality metrics stabilizes.
We compare the Kronecker method with other low rank methods and the baseline model in terms of number of parameters, running time and prediction
quality. We perform experiments about model compression based on implementation of the Kronecker layers in
Theano \cite{bergstra+al:2010-scipy,Bastien-Theano-2012} framework, and experiments about model speedup are based on Caffe \cite{jia2014caffe} framework.

\subsection{SVHN digits}
The SVHN dataset \cite{netzer2011reading} is a real-world digit recognition dataset consisting of photos of house numbers in Google Street View images.
Here we consider the digit recognition task where the inputs are 32-by-32 colored images centered around a single character.
There are 73257 digits for training, 26032 digits for testing, and 531131 less difficult samples which can be used as extra training data.
To build a validation set, we randomly select 400 images per class from training set and 200 images per class from extra training set as \cite{sermanet2012convolutional,DBLP:conf/icml/GoodfellowWMCB13} did.

Our baseline model has $8$ layers and the first $6$ layers consist of four convolutional layers and two pooling layers. The 7th layer is the
fully-connected layer and the 8th is the softmax output. The input of the fully-connected layer is of size $256\times5\times5$. The baseline's
fully-connected layer has $256$ hidden units.


Test results are listed in Table \ref{tab:svhn_approximation}. All results are averaged by $5$ models. In the SVD-$r$ methods
\cite{xue2013restructuring,denton2014exploiting}, we apply singular value decomposition on baseline's weight matrix, reconstruct it to rank $r$ and
fine-tune the restructured model. In the KFC-shape method, we replace the fully-connected layer by KFC layer with combination of $3$ different
formulations discussed above. $m_1 = 64, m_2 = 4$ in Formulation I, $m_1 = 128, m_2 = 2$ in Formulation II and Formulation III. In the KFC-rank
method, we replace the fully-connected layer by KFC layer with configuration $(m_1,m_2,n_1,n_2) = (64, 4, 256, 25, 5)$. Both KFC-shape and KFC-rank use additional nonlinearity and do not share bias.
\begin{table}[!ht] \centering \small
  \caption{Comparison of SVD method and KFC layers on SVHN digit recognition.}
\centering
\begin{tabular}{p{1.45cm} p{1.95cm} p{2cm} p{1.3cm}}
    \toprule \parbox[c]{\hsize}{Methods}  & \parbox[c]{1.95cm}{\#Layer Params\\(Reduction)} & \parbox[c]{2cm}{\#Model Params\\(Reduction)} & \parbox[c]{\hsize}{Test Error} \\
    \midrule Baseline & 1.64M & 2.20M & 2.57\% \\
    \midrule SVD-128 & 0.85M(2.0$\times$) & 1.42M(1.6$\times$) & 2.75\% \\
    \midrule SVD-64 & 0.43M(3.8$\times$) & 0.99M(2.2$\times$) & 2.85\% \\
    \midrule SVD-12 & 0.08M(20.0$\times$) & 0.65M(3.4$\times$) & 3.36\% \\
    \midrule KFC-shape & 0.34M(4.8$\times$) & 0.91M(2.4$\times$) & \textbf{2.60\%} \\
    \midrule KFC-rank & 0.08M(20.0$\times$) & 0.65M(3.4$\times$) & 2.82\% \\
\hline
\end{tabular}
\label{tab:svhn_approximation}
\end{table}

From the results we can see on SVHN digit recognition,
the KFC layer can reduce the number of parameters by $20\times$ with $0.25\%$ accuracy loss,
while SVD method will incur $0.79\%$ accuracy loss at the same compression ratio.

\subsection{SVHN sequences}
We also tested our models on SVHN digit sequence recognition.
Following the experimental setting as in~\cite{Ref:Goodfellow2013,Ref:Ba2015},
we preprocessed the training data by expanding the bounding box of each digit sequence by 30\% and resize the patch to $64\times64$ input.
Our model is built on top of the strong baseline CNN used in \cite{Ref:Jaderberg2015},
which by itself gives a 4.0\% whole-sequence error.
The baseline model has three large fully-connected layers, each with 3072 hidden units.
Based on this model, we replaced the first two fully-connected layers by two KFC layers of configuration
$ (m_1, m_2, n_1, n_2, r) = (60, 40 ,192, 64, 10) \text{ and } (30, 10, 60, 40, 10)$, and the third fully-connected
layers in the baseline model by a fully-connected layer with 300 hidden units.
The five parallel fully-connected layer for classification, with a total of 55 hidden units, is replaced by a fully-connected
layer of 220 hidden units followed by a maxout layer of 4 units.

This model achieves a 3.4\% sequence error rate, with only about 12\% parameters compared to the baseline CNN in \cite{Ref:Jaderberg2015}.
The performance of other competing methods are listed in Tab.~\ref{Tab:SVHNseq}.

\begin{table}[!ht]\centering \small
  \caption{Comparison of different methods on SVHN sequence recognition.}
\label{Tab:SVHNseq}
\begin{tabular}{p{6cm} p{1.5cm}}
\toprule \textbf{Model}                           & \textbf{Error Rate} \\
\midrule KFC                           & \textbf{3.4\%}      \\
\midrule ST-CNN (Multi)~\cite{Ref:Jaderberg2015} & 3.6\%               \\
\midrule ST-CNN (Single)~\cite{Ref:Jaderberg2015}  & 3.7\%               \\
\midrule DRAM~\cite{Ref:Ba2015}                   & 3.9\%               \\
\midrule CNN~\cite{Ref:Jaderberg2015}             & 4.0\%               \\
\midrule Maxout CNN~\cite{Ref:Goodfellow2013}     & 4.0\%               \\
\hline
\end{tabular}
\end{table}

\subsection{CASIA-HWDB}
CASIA-HWDB~\cite{Ref:liu2011casia} is an offline Chinese handwriting dataset, containing more than 3 million
character samples of over 7,000 character classes, produced by over 1,000 writers.
On this dataset, the best reported error rate we know of is from \cite{Ref:hwdb-googlenet}. Their model
uses a similar architecture of GoogLeNet \cite{Ref:googlenet} named HCCR-GoogLeNet, which reaches the single-model error rate of 3.74\%.

It is worth pointing out that in both GoogLeNet and HCCR-GoogLeNet,
a pooling layer with a considerable large size (7 or 5) is used to downsample the large feature maps generated by the inception layers, as
an approach to reduce the number of parameters in the next fully-connected layer.
Our KFC model, however, can directly operate on the large feature maps due to its nature of model compression.
Our architecture is based on HCCR-GoogLeNet, with the layers after the four inception groups replaced by
a KFC layer with configuration $ (m_1, m_2, n_1, n_2, r) = (128, 32, 832, 36, 5)$,
followed by two fully-connected layer with 1024 and 512 hidden units, respectively.
This model achieves an error rate of 3.37\%, which advances the previous state-of-the-art.
Although the original approach containing a large pooling layer actually uses fewer parameters, but using a large downsampling operation
inevitably loses much information and hence doesn't perform well enough.
Other competing methods are listed in Tab. ~\ref{tab:hwdb}.


\begin{table}[!ht]\centering \small
  \caption{Recognition Performances of different methods on CASIA-HWDB.}
\label{tab:hwdb}
\begin{tabular}{p{6cm} p{1.5cm}}
\toprule \textbf{Model}                           & \textbf{Error Rate} \\
\midrule KFC                           & \textbf{3.37\%}      \\
\midrule HCCR-GoogLeNet~\cite{Ref:hwdb-googlenet} & 3.74\%               \\
\midrule ATR-CNN Voting \cite{wu2014handwritten} & 3.94\%\\
\midrule CNN-Fujitsu ~\cite{yin2013icdar} & 5.23\%\\
\midrule MCDNN-INSIA \cite{yin2013icdar} & 5.58\%\\
\hline
\end{tabular}
\end{table}

\subsection{Scene Text Characters}
We use CNN described in \cite{jaderberg2014deep} to test our KConv layers since \cite{DBLP:conf/bmvc/JaderbergVZ14,lebedev2014speeding} both
experimented on this dataset. The dataset contains 186k $24\times 24$ character images cropped from some character recognition datasets. The network
has $4$ convolutional layers, a fully-connected layer and a $36$ classes softmax. The network uses maxout \cite{DBLP:conf/icml/GoodfellowWMCB13} as nonlinearity function after each convolutional layer and fully-connected layer. Training settings are almost the same as SVHN.
We test the model speed on Nvidia GPU with Caffe \cite{jia2014caffe} framework.

We replace the second and third convolutions by our KConv layer since these two layers constitute more than 90\% of the running time. The second convolution has $48$ input channels and $128$ output channels with $9\times 9$ filters. The third convolution has $64$ input channels and $512$ output channels and filters of size $8\times 8$.

The results are shown in Table \ref{tab:kconv}. The KConv layer can achieve about $3.3\times$ speedup on the whole model with less than 1\% accuracy loss. The result is similar to \cite{DBLP:conf/bmvc/JaderbergVZ14}, as the KConv layer includes the Jaderberg-style rank-1 filter as a special case. The reported results are measured by validation errors as we are only concerned with relative performance of each method.

\begin{table}[!ht] \centering \small
\caption{Speedup of KConv on scene text character recognition dataset. Parameters of the different layers are separated by a semicolon.}
 \centering
\begin{tabular}{p{1.5cm} p{2.3cm} p{1.6cm} p{1.3cm}}
    \toprule Methods  & \parbox[]{2.3cm}{Configuration\\($r$,$o_1$,$c_1$,$h_1$,$w_1$)} & Validation Error & Speedup\\
    \midrule Baseline & / &7.84\% & / \\
    \midrule \parbox[c]{\hsize}{KConv-a} & \parbox[]{2.3cm}{1,128,24,9,1;\\ 1,256,64,8,1} &8.76\% & 3.3$\times$ \\
    \midrule \parbox[c]{\hsize}{KConv-b} & \parbox[]{2.3cm}{1,128,48,1,9;\\ 1,512,64,1,8} &8.69\% & 3.0$\times$ \\
    \midrule \parbox[c]{\hsize}{KConv-c} & \parbox[]{2.3cm}{2,64,24,9,1;\\ 2,256,64,8,1} &7.87\% & 2.9$\times$ \\
\hline
\end{tabular}
\label{tab:kconv}
\end{table}

We have also experimented replacing the first convolutional layer with KConv layer. In this case, KConv with ($r$,$o_1$,$c_1$,$h_1$,$w_1$) = (2, 12, 1, 1, 9), is found to outperform Jaderberg-style rank-1 filter with ($r$,$o_1$,$c_1$,$h_1$,$w_1$) = (2, 96, 1, 1, 9) by 0.83\%.

\subsection{Scene Text Words}
We also experiment on the word recognition model trained on the synthetic word dataset consisting of 9 million images covering about 90k English words from \cite{Jaderberg14c} and tested on ICDAR 2013\cite{karatzas2013icdar} dataset.
As the model predicts English word, the number of output classes is about 90k, resulting in a model with more than 400 million parameters,
mostly in the fully-connected layers.

\begin{figure}
\centering
\includegraphics[height=0.52 \linewidth, width= 0.9\linewidth]{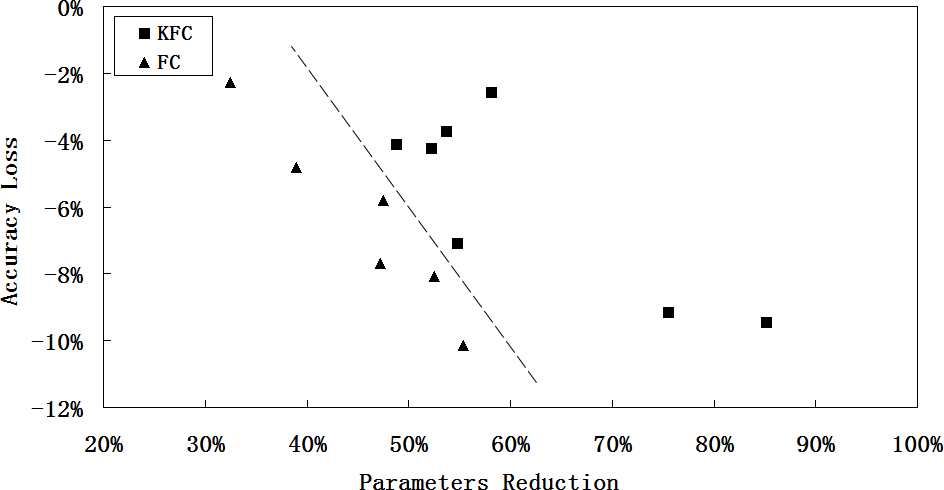}
\caption{Accuracy loss and total parameter reduction on ICDAR'13 with different models. The KFC indicates models using the KFC layer and the FC indicates models without the KFC layer.}
\label{fig:90k}
\end{figure}

We use
different shapes and ranks in experiments of the KFC layers to replace the last FC layer.
For comparison, we test a method which simply decrease the number of output neurons of
the second-to-last FC layer before softmax. We also test the method in \cite{sainath2013low} which also tries to approximate the last weight matrix. Figure \ref{fig:90k}
list the test results. Due to lack of space, we have not listed the detailed hyper-parameters of all these experiments.
The KFC model with highest
accuracy uses a configuration with shapes (26, 15, 719, 122), (26, 15, 122, 719), (13, 30, 61, 1438), (130, 3, 1438, 61),
each shape of rank $10$,
building a KFC layer of total rank 40. But this layer itself still saves 92\% parameters compared to its FC counterpart.
The scatter diagram indicates that the KFC layer requires less parameter with the same accuracy or has higher accuracy with the same number of
parameters. This demonstrates that our technique also works well before softmax layer.

\subsection{ImageNet}
ImageNet (ILSVRC12) \cite{ILSVRC15} is a large scale visual recognition dataset and contains 1000 categories and 1.2 million images for training. We
use the AlexNet \cite{krizhevsky2012imagenet} as the baseline network and use the implementation in \cite{ding2014theano}. The AlexNet has three
fully-connected layers. The first's input is a tensor of size $256\times 6 \times 6$ and the weight matrix is of size $9216\times 4096$. The second
and the third layers have weight matrices with size $4096\times 4096$ and $4096 \times 1000$, respectively.
\begin{table}[!ht] \centering \small
\caption{Comparison of using SVD method and using KFC layers on ImageNet.}
\centering
\begin{tabular}{p{1.5cm} p{2cm} p{1.6cm} p{1.6cm}}
    \toprule \parbox[c]{\hsize}{Methods} & \parbox[]{2cm}{\#Model Params\\(Reduction)} & Top 1 Error & Top 5 Error\\
    \midrule Baseline& / & 42.61\% & 19.86\% \\
    \hline
    \midrule SVD-2 & 16.9M(3.6$\times$) & 43.80\% & 20.78\% \\
    \midrule KFC-2 & 16.9M(3.6$\times$) & 43.01\% & \textbf{20.05\%} \\
    \hline
    \midrule SVD-3 & 6.1M(10.0$\times$) & 45.67\% & 21.52\% \\
    \midrule KFC-3 & 6.1M(10.0$\times$) & 45.33\% & \textbf{21.10\%} \\
\hline
\end{tabular}
\label{tab:imagenet}
\end{table}
Test results are listed in Table~\ref{tab:imagenet}. SVD-2 and KFC-2 compress the first two fully-connected layers, and SVD-3 and
KFC-3 compress all the three fully-connected layers. We select the hyper-parameters carefully to ensure that the two comparing methods have the same model size. In SVD-2, the ranks are $237$ and $1000$. In SVD-3, the ranks are $100$, $165$, $200$. In KFC-2,
the configurations are (1024, 4, 1536, 6, 2) for the first layer and (2048, 2, 2048, 2, 2) for the second.
In KFC-3 the configurations are (512, 8, 512, 18, 5), (512, 8, 512, 8, 5) and (500, 2, 512, 8, 4). We use additional nonlinearity in all
KFC layers. The results demonstrate that we can look for the best or most suitable choice from a variety of different shapes and ranks in KFC layers.

\section{Related Work}
In this section we discuss some related works not yet covered.

In addition to the low rank methods mentioned earlier, hashing methods have also been used to reduce the number of parameters \cite{DBLP:conf/icml/ChenWTWC15,bakhtiary2015speeding}, and distillation offers another way of compressing neural networks \cite{hinton2015distilling}. Furthermore, \citeauthor{mathieu2013fast} (\citeyear{mathieu2013fast}) used FFT to speedup convolution. \citeauthor{DBLP:conf/aistats/YangWSS15} (\citeyear{DBLP:conf/aistats/YangWSS15}) used adaptive fastfood transform to reparameterize the matrix-vector multiplication of fully-connected layers.  \citeauthor{han2015learning} (\citeyear{han2015learning}) iteratively pruned redundant connection to reduce the number of parameters. \citeauthor{gong2014compressing} (\citeyear{gong2014compressing}) used vector quantization to compress the fully-connected layer.  \citeauthor{DBLP:conf/icml/GuptaAGN15} (\citeyear{DBLP:conf/icml/GuptaAGN15}) suggested using low precision arithmetic to compress the neural network.

\section{Conclusion}
In this paper, we have proposed and studied a framework to reduce the number of parameters and computation time in convolutional neural networks. Our
framework uses Kronecker products to exploit the local structure within convolutional layers and fully-connected layers. As Kronecker product is a
generalization of the outer product, our method generalizes the low rank approximation method for matrices and tensors. We also explored combining
Kronecker product of different shapes to further balance the drop in accuracy and the reduction in parameters. Method for initializing Kronecker layer is also given.

Through a series of experiments on different datasets, our method is proven to be effective and efficient on different tasks.
It can reduce the computation time and model size with minor lost in accuracy, or improve previous state-of-the-art performance with similar model size.

A key advantage of our method is that Kronecker layers can be implemented by combination of tensor reshaping operation and dense matrix product, which can be efficiently performed on CPU. The generality of Kronecker product also allows a lot of freedom to trade off between model size, running time and prediction accuracy through the selection of the hyper-parameters in Kronecker layers such as shapes and ranks.


\bibliographystyle{aaai}
\fontsize{9.5pt}{10.5pt} \selectfont
\bibliography{paper}

\end{document}